\documentclass{article}
\usepackage{ijcai21}

\usepackage{times}
\usepackage{soul}
\usepackage{url}
\usepackage[hidelinks]{hyperref}
\usepackage[utf8]{inputenc}
\usepackage[small]{caption}
\usepackage{graphicx}
\usepackage{amsmath}
\usepackage{amsthm}
\usepackage{booktabs}
\usepackage{algorithm}
\usepackage{algorit hmic}
\newcommand{\citet}[1] {\citeauthor{#1}~\shortcite{#1}}

\title{Document-level Relation Extraction as Semantic Segmentation}

\author{
Ningyu Zhang\textsuperscript{\rm 1,2 \thanks{ \quad Equal contribution and shared co-first authorship.} } \and 
Xiang Chen \textsuperscript{\rm 1,2 \footnotemark[1]} \and
\textbf{Xin Xie}\textsuperscript{\rm 1,2}  \and
\textbf{Shumin Deng}\textsuperscript{\rm 1,2} \and
\textbf{Chuanqi Tan}\textsuperscript{\rm 3} \and \\
\textbf{Mosha Chen}\textsuperscript{\rm 3} \and 
\textbf{Fei Huang}\textsuperscript{\rm 3} \and
\textbf{Luo Si}\textsuperscript{\rm 3} \and
\textbf{Huajun Chen}\textsuperscript{\rm 1,2 \thanks{\quad Corresponding author.}} \\
\affiliations

\textsuperscript{\rm 1} Zhejiang University \& AZFT Joint Lab for Knowledge Engine \\
\textsuperscript{\rm 2} 
Hangzhou Innovation Center, Zhejiang University \\
\textsuperscript{\rm 3} Alibaba Group\\

\emails
 \{zhangningyu,xiang\_chen,xx2020,231sm,huajunsir\}@zju.edu.cn \\
 \{chuanqi.tcq,chenmosha.cms,f.huang,luo.si\}@alibaba-inc.com \\
}

\begin{document}

\maketitle

\begin{abstract}
Document-level relation extraction aims to extract relations among multiple entity pairs from a document. Previously proposed graph-based or transformer-based models utilize the entities independently, regardless of global information among relational triples. This paper approaches the problem by predicting an entity-level relation matrix to capture local and global information, parallel to the semantic segmentation task in computer vision. Herein, we propose a Document U-shaped Network for document-level relation extraction. Specifically, we leverage an encoder module to capture the context information of entities and a U-shaped segmentation module over the image-style feature map to capture global interdependency among triples. Experimental results show that our approach can obtain state-of-the-art performance on three benchmark datasets DocRED, CDR, and GDA\footnote{The code and datasets are available in \url{https://github.com/zjunlp/DocuNet}.}.
\end{abstract}

\section{Introduction} 

Relation extraction (RE) is an important task in the field of information extraction, which has widespread applications \cite{zhang2021cause,zhang2021alicg}. Previous works \cite{zeng2015distant,feng2018reinforcement} focused on identifying relations within a single sentence, which failed to recognize relations between entities across sentences.
However, many relations are expressed over multiple sentences in real-world applications. According to \cite{Yao2019DocREDAL}, above 40.7\% of relations can only be identified at the document level. Therefore, it is crucial for models to be able to extract document-level relations. 
\begin{figure}[h]
  \centering 
  \includegraphics[width=0.45\textwidth]{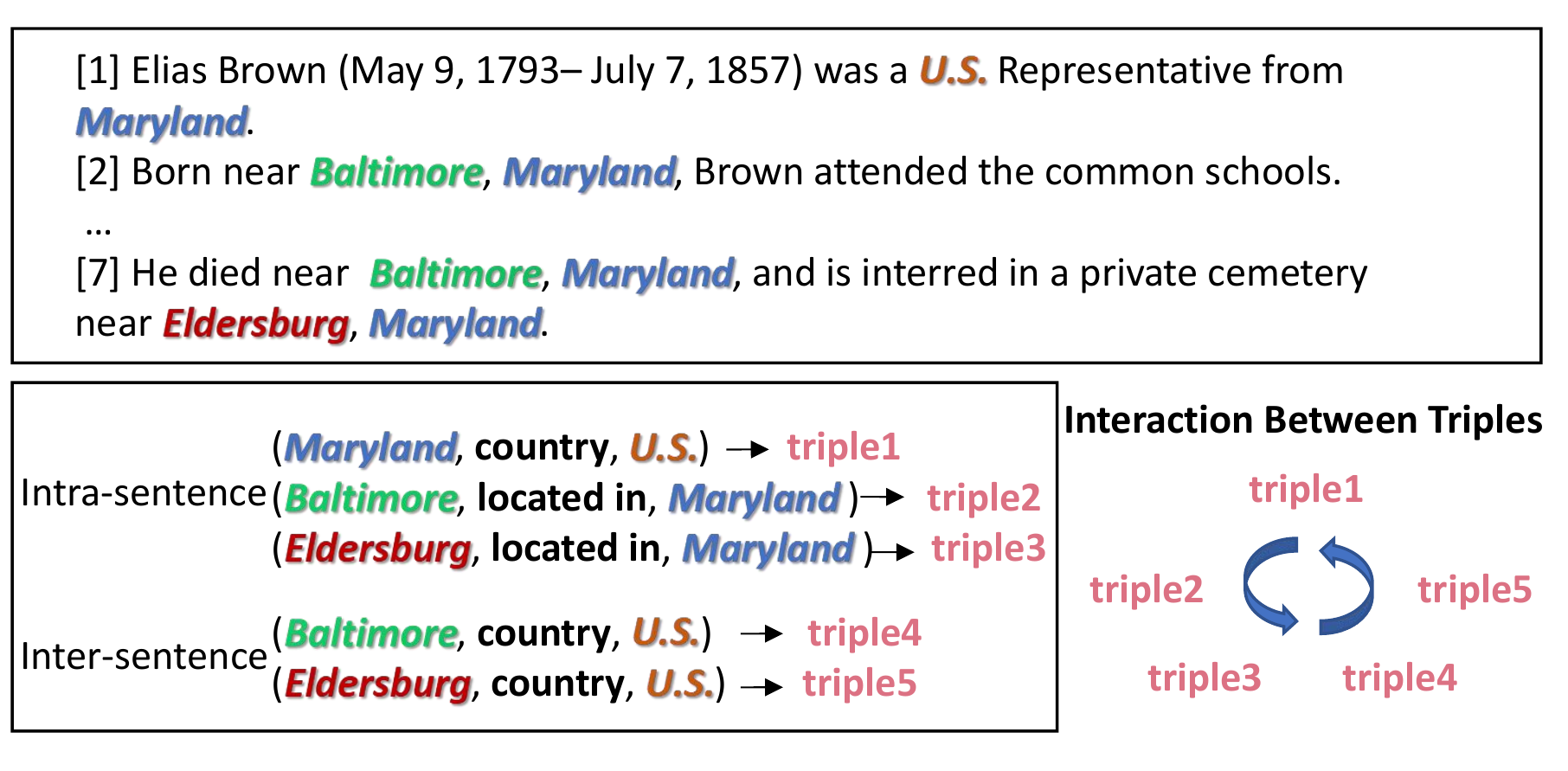} 
  \caption{Example document with entity pairs and relations from DocRED. Entity mentions and relations only involved in these relation instances are colored.}
 \label{task}
\end{figure}

Recent studies \cite{Yao2019DocREDAL,tang2020hin,zeng2020double,wang2020global,zhou2020document} have extended sentence-level RE to the document level. Compared with sentence-level RE that only contains one entity pair to classify in a sentence, document-level RE requires the model to classify the relations of multiple entity pairs at once.
Besides, the subject and object entities involved in a relation may appear in different sentences. Therefore a relation cannot be identified based solely on a single sentence. 
For example, as shown in Figure \ref{task}, it is easy to identify the intra-sentence relations, such as (\emph{Maryland}, country, \emph{U.S.}), (\emph{Baltimore}, located\_in, \emph{Maryland}), and (\emph{Eldersburg}, located\_in, \emph{Maryland}), owing to the occurrence of entities in the same sentence. However, it is more challenging for a model to recognize inter-sentence relations, such as those between \emph{Eldersburg} and \emph{U.S.} and between \emph{Baltimore}  and \emph{U.S.} because these mentions occur in different sentences and have long-distance dependencies. 

To extract relations among these inter-sentence entity pairs, most current studies constructed document-level graph module based on heuristics, structured attention or dependency structures~\cite{Peng2017CrossSentenceNR,Christopoulou2019ConnectingTD,Nan2020ReasoningWL,zeng2020double,wang2020global}, followed by reasoning with graph neural models. Meanwhile, considering the transformer architecture can implicitly model long-distance dependencies, some studies \cite{Wang2019FinetuneBF,tang2020hin,zhou2020document} directly applied pre-trained language models rather than explicit graph reasoning. In general, current approaches obtain entity representation via information passing through nodes on document-level graphs or transformer-based structure learning. 
However, they mainly focus on token-level syntactic features or contextual information rather than global interactions between entity pairs, neglecting the interdependency among the multiple relations in one context.  

Concretely, the interdependency among multiple triples is advantageous and can provide guidance for relation classification in the case of many entities. For example, if the intra-sentence relation (\emph{Maryland}, country, \emph{U.S.}) has been identified, it is implausible for \emph{U.S.} to be in any other person-social relationship, such as "is the father of...".
Besides, according to the triples that \textit{Eldersburg} is located in \textit{Maryland} and \textit{Maryland} belongs to \textit{U.S.}, we can infer that \textit{Eldersburg} belongs to \textit{U.S.}.
As described above, each relation triple can provide information to other relation triples in the same text. 

\begin{figure}[h]
  \centering 
  \includegraphics[width=0.35\textwidth]{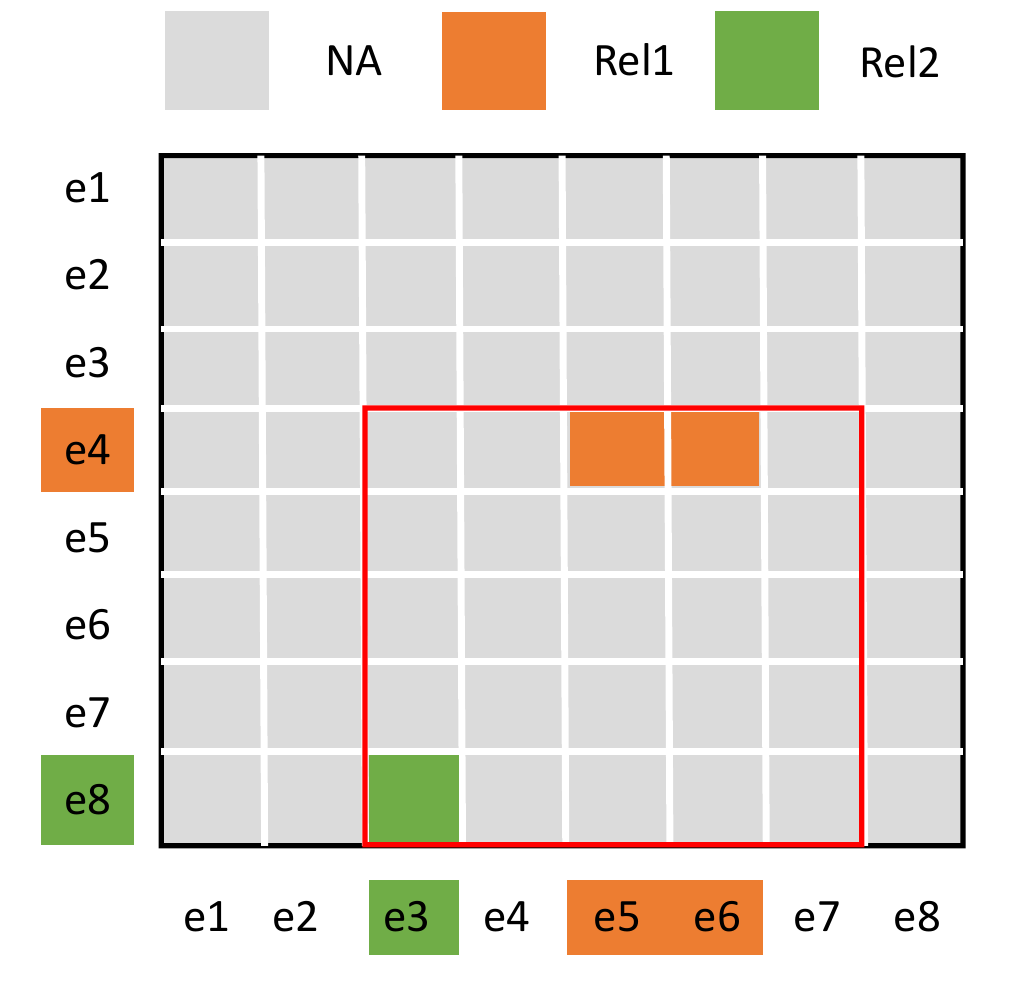} 
  \caption{Illustration of the entity-level relation matrix applied in our formulation. Each cell belongs to one relation type.}
  \label{motivation}
\end{figure}

To capture the interdependency among the multiple triples, we reformulate the document-level RE task as an entity-level classification problem \cite{jiang2019generalizing}, also known as table filling \cite{miwa2014modeling,gupta2016table}, as shown in Figure \ref{motivation}. It is analogous to semantic segmentation (a well-known computer vision task), whose goal is to label each pixel of the image with the corresponding represented class by convolution network.
Inspired by the above, we propose a novel model called \textbf{Doc}ument \textbf{U}-shaped \textbf{N}etwork (\textbf{DocuNet}), which formulates document-level RE as semantic segmentation. In this manner, given relevant features between entity pairs as an image, the model predicts the relation type for each entity pair as a pixel-level mask.  
Specifically, we introduce an encoder module to capture the context information of entities and a U-shaped segmentation module over the image-style feature map to capture global interdependency among triples. We further propose a balanced softmax method to handle the imbalance relation distribution. 
Our contributions can be summarized as follows:
\begin{itemize}
    \item To the best of our knowledge, this is the first approach that regards document-level RE as a semantic segmentation task. 
    \item We introduce the model DocuNet to capture both local context information and global interdependency among triples for document-level RE. 
    \item Experimental results on three benchmark datasets show that our model DocuNet can achieve state-of-the-art performance compared with baselines.
\end{itemize}

\section{Related Work}

Previous relation extraction approaches mainly concentrate on identifying the relation between two entities within a sentence. Many approaches~\cite{zeng2015distant,feng2018reinforcement,zhang2018attention,zhang2019long,zhang2020relation,DBLP:conf/emnlp/ZhangDBYYCHZC20,DBLP:journals/corr/abs-2009-09841,ye2020contrastive,yu2020bridging,DBLP:journals/corr/abs-2009-09841,wu2021curriculum,DBLP:journals/corr/abs-2104-07650,zheng2021prgc} have been proposed to tackle the sentence-level RE task effectively. However, sentence-level RE faces an inevitable restriction in that many real-world relations can only be extracted by reading multiple sentences. For this reason, document-level RE appeals to many researchers~\cite{tang2020hin,Nan2020ReasoningWL,zeng2020double,wang2020global,xiao2020denoising}.

Various approaches for document-level RE mainly include graph-based models and transformer-based models. Graph-based approaches are now widely adopted in RE because of their effectiveness and strength in relational reasoning.
\citet{Jia2019DocumentLevelNR} proposed a model that combines representations learned over various text spans throughout the document and across the sub-relation hierarchy.
\citet{Christopoulou2019ConnectingTD} proposed an edge-oriented graph neural model (EoG) for document-level RE.  
\citet{li2020graph} characterized the complex interaction between sentences and potential relation instances with a graph-enhanced dual attention network (GEDA).
\citet{zhang2020document} proposed a novel graph-based model with a Dual-tier Heterogeneous Graph (DHG), which contains a structure modeling layer followed by a relation reasoning layer. \citet{zhou2020global} proposes a global context-enhanced graph convolutional network (GCGCN), composed of entities as nodes and the contexts of entity pairs as edges between nodes.
\citet{wang2020global} proposed a novel model (GLRE) that encodes the document information in terms of global and local entity representations as well as context relation representations. 
\citet{Nan2020ReasoningWL} proposed a novel model (LSR) that enables relational reasoning across sentences by automatically inducing a latent document-level graph. 
\citet{zeng2020double} proposed the graph aggregation-and-inference network (GAIN) with double graphs for document-level RE.  
\citet{xu2020document} proposed an encoder-classifier reconstructor model (HeterGSAN), which manages to reconstruct the ground-truth path dependencies from the graph representation. 
Explicit graph reasoning can bridge the gap between entities that occur in different sentences, thus mitigating long-distance dependency and achieving promising performance.
 \begin{figure*}[h]
  \centering 
  \includegraphics[width=0.85\textwidth]{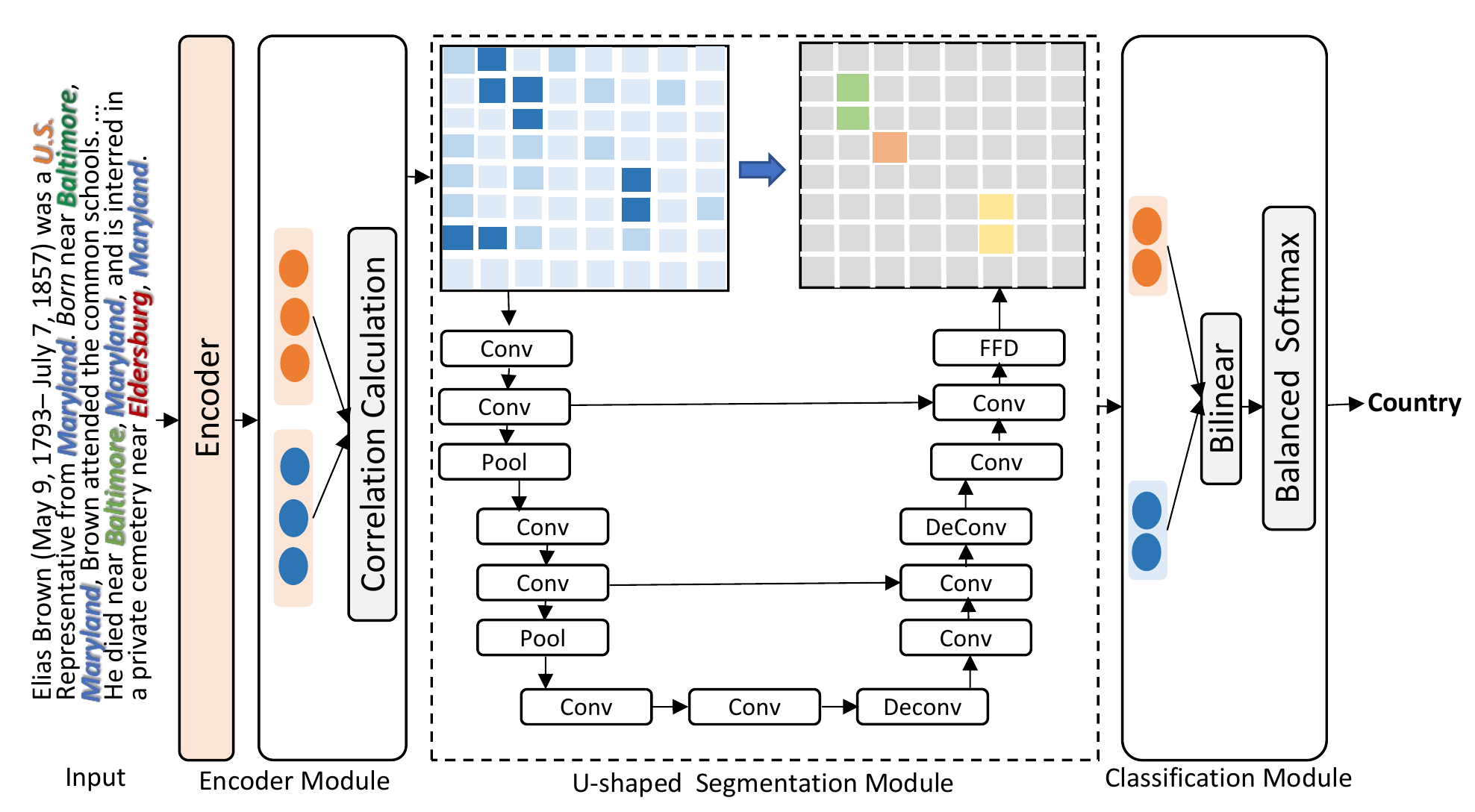} 
  \caption{Architecture of our \textbf{Doc}ument \textbf{U}-shaped \textbf{N}etwork (\textbf{DocuNet}) (Best viewed in color).}
  \label{arc}
\end{figure*}

In contrast, considering the transformer architecture can implicitly model long-distance dependencies, some researchers directly leverage pre-trained language models without generating document graphs.
 \citet{Wang2019FinetuneBF} proposed a two-step training paradigm on DocRED using BERT as pre-trained word embedding. They observed an imbalance in the distribution of relation and disentangled the relation identification and classification for better inference. 
 \citet{tang2020hin} proposed a hierarchical inference network (HIN) to make full use of the abundant information from the entity, sentence, and document levels to perform hierarchical reasoning. 
 \citet{zhou2020document} proposed a novel transformer-based model (ATLOP) of adaptive thresholding and localized context pooling based on BERT. However, most previous studies focused on the local entity representation, regardless of the high-level global connections between triples, which overlooked the interdependency between multiple relations.

On the one hand, our work is inspired by \cite{jin2020relation}, which was the first to consider the issue of global interaction between relations, and there have been few studies on RE. 
On the other hand, as these studies\cite{DBLP:conf/naacl/NguyenG15,DBLP:conf/coling/ShenH16} have done, convolutional neural networks have been long used in the relation extraction area, which enlightens us to pay attention to the role of CNN in extracting information of the image-style feature map.
Hence, our work is also related to the study of \cite{liu2020incomplete}, who formulated incomplete utterance rewriting as a semantic segmentation task and motivated us to study the RE problem from a computer vision perspective. 
In this study, we leveraged the U-Net \cite{RFB15a}, which consists of a contracting path to capture context and a symmetric expanding path that enables precise localization. To the best of our knowledge, this is the first approach to formulate RE as a  semantic segmentation task.

\section{Methodology}\label{sec:method_model}

\subsection{Preliminary}
We first introduce the problem definition. With a document $d$ containing a set of entities  $\{e_i\}_{i=1}^n$, the task is to extract the relations between entity pairs  $(e_s, e_o)$. In one document, each entity $e_i$ may occur multiple times.  To model relation extraction between $e_s$ and $e_o$, we define a $N\times N$ matrix $Y$, where entry $Y_{s,o}$ indicates the relation type between $e_s$ and $e_o$. Then, we obtain the output of matrix $Y$, analogous to the task of semantic segmentation. Entities in $Y$ are arranged according to their first appearance in the document. We obtain the feature map via the entity-to-entity relevance estimation and take the feature map as an image. Note that the output entity-level relation matrix $Y$ is parallel to the pixel-level mask in semantic segmentation, which bridges relation extraction and semantic segmentation. Our approach can also be applied to sentence-level relation extraction. Since the document has relatively more entities, thus, entity-level relation matrix can learn more global information to boost the performance.

\subsection{Encoder Module}
Given the document $d=[x_t]_{t=1}^L$, we insert special symbols "$<e>$" and "$</e>$" at the start and end of mentions to mark the entity positions. We leverage the pre-trained language model as an encoder to obtain the embedding as follows:
\begin{equation}
    {H}=\left[{h}_1, {h}_2, ..., {h}_L\right] = \text{Encoder}(\left[x_1, x_2, ..., x_L\right]). \label{eq:encoder}
\end{equation}
where $h_i$ is the embedding of the token $x_i$. Note that some documents are longer than 512, we thus leverage a \emph{dynamic window} to encode whole documents. We average the embeddings of overlapping tokens of different windows to obtain the final representations. Then, we utilize the embeddings of "$<e>$" to represent mention following~ \cite{Verga2018SimultaneouslyST}. We leverage a smooth version of max pooling, namely, logsumexp pooling~\cite{Jia2019DocumentLevelNR} each entity  $e_i$, to obtain the entity embedding $\mathbf{e}_{i}$:
\begin{equation}
    \mathbf{e}_{i} = \log \sum_{j=1}^{N_{e_i}} \exp \left( \mathbf{m_j} \right). \label{eq:pool}
\end{equation}
This pooling accumulates signals from mentions in the document. Thus, we obtain the entity embedding $\mathbf{e}_{i}$.

We calculate the entity-level relation matrix based on entity-to-entity relevance. For each entity $e_i$ in the matrix, their relevance is captured by a $D$-dimensional feature vector $\mathbf{F}(e_s, e_o)$. We introduce two strategies for computing $\mathbf{F}(e_s, e_o)$, namely, \emph{similarity-based} method and \emph{context-based} method. Similarity-based method is produced by concatenating operation result of element-wise similarity, cosine similarity and bi-linear similarity between $e_s$ and $e_o$ as:

\begin{equation}
    \! \mathbf{F}(e_s, e_o) \!=\! \big[ e_s\!\odot e_o; cos({ e_s,\! e_o)}; e_s W_1 e_o    \big]\!,\!\!\!\!  
\end{equation}
For the context-based strategy, we leverage entity-aware attention with affine transformation to obtain the feature vector as follows:
\begin{equation}
    \! \mathbf{F}(e_s, e_o) \!= W_2 H a^{(s,o)} 
\end{equation}
 \begin{equation}
     a^{(s,o)} = softmax(\sum_{i=1}^{K} A_{i}^s \cdot A_{i}^o)
 \end{equation}
 where $a^{(s,o)}$ is the attention weight for entity-aware attention and $A_{i}^s$ refers to the tokens' importance to the $i$-th entity, $H$ is the document embedding, $W_1$, $W_2$ is the learnable weight matrix, $K$ is the number of head in the transformer. 

\subsection{U-shaped Segmentation Module}
Taking the entity-level relation  matrix $\mathbf{F}\,{\in}\,R^{N{\times}N{\times}D}$ as a $D$-channel image, we formulate the document-level relation prediction  as the pixel-level mask in $F$. where $N$ is the largest number of entities, counted from all the dataset samples. Specifically $N$ is the largest number of entities, counted from all the dataset samples. To this end, we utilize U-Net~\cite{RFB15a}, which is a famous semantic segmentation model in computer vision. 
As can be seen in Figure \ref{arc},
the module is formed as a U-shaped segmentation structure, which contains two down-sampling blocks and two up-sampling blocks with skip connections. On the one hand, each down-sampling block has two subsequent max pooling and separate convolution modules. Further, the number of channels is doubled in each down-sampling block.
As it shows in the Figure \ref{motivation}, 
the segmentation area in the entity-level relation matrix refers to the co-occurrence of relations between entity pairs. The U-shaped segmentation structure can promote the information exchange between entity pairs in the receptive field analogy to implicit reasoning. Specifically, CNN and down-sampling block can enlarge the receptive field of current entity pair embedding $\mathbf{F}(e_s, e_o)$, thus, providing rich global information for representation learning. 
On the other hand, the model has two up-sampling blocks with a subsequent deconvolution neural network and two separate convolution modules. Different from down-sampling, the number of channels is halved in each up-sampling block, which can distribute the aggregated information to each pixel.

Finally, we incorporate an encoding module and a U-shaped segmentation module to capture both local and global information $Y$ as follows:
\begin{equation}
    \! \mathbf{Y} \!= U(W_3 \mathbf{F}) 
\end{equation}
where $U$ and $\mathbf{Y}\,{\in}\,R^{N{\times}N{\times}D'}$  denote the U-shaped segmentation module and entity-level relation matrix respectively. $W_3$ is the learnable weight matrix in order to reduce the dimension of $F$ and $D'$ is much smaller than $D$.

\subsection{Classification Module}
Given the entity pair embedding $\textbf{e}_s$ and $\textbf{e}_o$ with the entity-level relation matrix $Y$, we map them to hidden representations $z$ with a feedforward neural network. Then, we obtain the probability of relation via a bilinear function. Formally, we have: 
\begin{equation}
    {z}_s =\tanh \left( {W}_s \mathbf{e_s} + Y_{s,o}\right), \label{eq:t1}
   \end{equation}
   \begin{equation}
    {z}_o =\tanh \left( {W}_o \mathbf{e_o} + Y_{s,o}\right), \label{eq:t2}
       \end{equation}
\begin{equation}
    \mathrm{P}\left(r|e_s, e_o\right) = \sigma \left({z}_s  {W}_r {z}_o  + b_r\right),
\end{equation}

where $Y_{s,o}$ is the entity-pair representation of $(s,o)$ in matrix $Y$,
${W}_r\in  R^{d \times d}$, $b_r \in  R$, ${W}_s \in  R^{d \times d}$, and ${W}_o \in  R^{d \times d}$, are learnable parameters.
  
Since previous work \cite{Wang2019FinetuneBF} observed that there is an imbalance relation distribution for RE (many entity pairs have relation of \emph{NA}), we introduce a balanced softmax method for training, which is inspired by the circle loss \cite{sun2020circle} from computer vision. Specifically, we introduce an additional category $0$, hoping that the scores of the target category are all greater than $s_0$ and the scores of the non-target categories are all less than $s_0$. Formally, we have \cite{jianlinsu}:
\begin{equation}
L = \log \left(e^{s_{0}}+\sum_{i \in \Omega_{n e g}} e^{s_{i}}\right)+\log \left(e^{-s_{0}}+\sum_{j \in \Omega_{p o s}} e^{-s_{j}}\right).
\end{equation}

For simplicity, we set the threshold as zero and have the following: 

\begin{equation}
   L  =  \log \left(1+\sum_{i \in \Omega_{n e g}} e^{s_{i}}\right)+\log \left(1+\sum_{j \in \Omega_{p o s}} e^{-s_{j}}\right).
\end{equation}

\section{Experiments}

\subsection{Dataset}
\begin{table}[!t]
\centering
\scalebox{0.95}{
    \begin{tabular}{p{4cm}ccc}
         \toprule
         \textbf{Statistics / Dataset} & \textbf{DocRED} & \textbf{CDR} & \textbf{GDA} \\
         \midrule
         \# Train& 3,053& 500&  23,353 \\
         \# Dev& 1,000& 500& 5,839 \\
         \# Test& 1,000& 500& 1,000 \\
         \# Relations& 97& 2& 2 \\
         Avg. \# entities per Doc.& 19.5& 7.6& 5.4\\
        Avg. \# Ment. per Ent.& 1.4& 2.7& 3.3\\
         \bottomrule
    \end{tabular}
    }
    \caption{Statistics of the experimental datasets.}
    \label{tab::statistics}
\end{table}

\begin{table*}[htbp]
\centering
    \begin{tabular}{p{6.5cm}cccc}
         \toprule
         \textbf{Model} & \multicolumn{2}{c}{\textbf{Dev}} & \multicolumn{2}{c}{\textbf{Test}} \\
          & Ign $F_1$ & $F_1$ & Ign $F_1$ & $F_1$ \\
         \midrule
         GEDA-BERT$_{\text{base}}$~\cite{li2020graph}& 54.52& 56.16 & 53.71& 55.74 \\
         LSR-BERT$_{\text{base}}$~\cite{Nan2020ReasoningWL}& 52.43& 59.00& 56.97& 59.05\\
         GLRE-BERT$_{\text{base}}$~\cite{wang2020global}& - & - & 55.40 &57.40\\
         GAIN-BERT$_{\text{BASE}}$~\cite{zeng2020double}& 59.14&61.22& 59.00& 61.24 \\
         HeterGSAN-BERT$_{\text{base}}$~\cite{xu2020document}&58.13 &60.18 &57.12  &59.45   \\
         \midrule
         BERT$_{\text{base}}$~\cite{Wang2019FinetuneBF}& -& 54.16& -& 53.20 \\
         BERT-TS$_{\text{base}}$~\cite{Wang2019FinetuneBF}& -& 54.42&-& 53.92 \\

         HIN-BERT$_{\text{base}}$~\cite{tang2020hin}& 54.29& 56.31& 53.70& 55.60 \\
         CorefBERT$_{\text{base}}$~\cite{Ye2020CoreferentialRL}& 55.32& 57.51& 54.54& 56.96 \\
         ATLOP-BERT$_{\text{base}}$~\cite{zhou2020document}& 59.22 & 61.09& 59.31& 61.30 \\
         \midrule
         \midrule
         DocuNet-BERT$_{\text{base}}$&\textbf{59.86}$\pm$\textbf{0.13}&\textbf{61.83}$\pm$\textbf{0.19} & \textbf{59.93}& \textbf{61.86} \\
         \midrule
         BERT$_{\text{large}}$~\cite{Ye2020CoreferentialRL} & 56.67  & 58.83 & 56.47 & 58.69\\
         CorefBERT$_{\text{large}}$~\cite{Ye2020CoreferentialRL}& 56.82  & 59.01 & 56.40 & 58.83\\
         RoBERTa$_{\text{large}}$~\cite{Ye2020CoreferentialRL}  & 57.14 & 59.22 & 57.51 & 59.62 \\
         CorefRoBERTa$_{\text{large}}$~\cite{Ye2020CoreferentialRL}& 57.35  & 59.43 & 57.90 & 60.25\\
         ATLOP-RoBERTa$_{\text{large}}$~\cite{zhou2020document}& 61.32 & 63.18& 61.39& 63.40 \\
         \midrule
         DocuNet-RoBERTa$_{\text{large}}$&\textbf{62.23}$\pm$\textbf{0.12} &\textbf{64.12}$\pm$\textbf{0.14} & \textbf{62.39} & \textbf{64.55} \\
         \bottomrule
    \end{tabular}
    \caption{Results (\%) on the development and test set of DocRED. We run experiments five times with different random seeds and report the mean and standard deviation on the development set. We report the official test score on the CodaLab scoreboard with the best checkpoint on the development set.}
    \label{tab::main_results}
\end{table*}

\begin{table}[htbp]
\centering
    \begin{tabular}{p{4.5cm}cc}
         \toprule
         \textbf{Model} & CDR& GDA \\
         \midrule
         BRAN~\cite{Verga2018SimultaneouslyST}& 62.1& - \\
         EoG~\cite{Christopoulou2019ConnectingTD}& 63.6& 81.5 \\
         LSR~\cite{Nan2020ReasoningWL}& 64.8& 82.2\\
         DHG~\cite{zhang2020document}& 65.9 & 83.1 \\
         GLRE~\cite{wang2020global}& 68.5 & - \\
         SciBERT$_{\text{base}}$~\cite{Beltagy2019SciBERTAP}& 65.1 & 82.5 \\
         ATLOP-SciBERT$_{\text{base}}$~\cite{zhou2020document}&  69.4 &  83.9 \\
         \midrule
         DocuNet-SciBERT$_{\text{base}}$& \textbf{76.3$\pm$0.40}  & \textbf{85.3$\pm$0.50}  \\
         \bottomrule
    \end{tabular}
    \caption{Results (\%) on the biomedical datasets CDR and GDA.}
    \label{tab::bio_result}
\end{table}
 
 We evaluated our DocuNet model on three document-level RE datasets. We listed the dataset statistics in Table~\ref{tab::statistics}.
\begin{itemize}
    \item \textbf{DocRED}~\cite{Yao2019DocREDAL} is a large-scale document-level relation extraction dataset by crowdsourcing. DocRED contains 3,053/1,000/1,000 instances for training, validating and test, respectively.
    
    \item \textbf{CDR}~\cite{Li2016BioCreativeVC} is a relation extraction dataset in the biomedical domain, which is aimed to infer the interactions between chemical and disease concepts.
   
    \item \textbf{GDA}~\cite{Wu2019RENETAD} is a dataset in the biomedical domain, which consists of 23,353 training samples. Differently, the dataset is aimed to predict the interactions between disease concepts and genes.  
\end{itemize}
\subsection{Experimental Settings}
Our model was implemented based on Pytorch.
We used cased BERT-base, or RoBERTa-large as the encoder on DocRED and SciBERT-base~\cite{Beltagy2019SciBERTAP} on CDR and GDA. We optimize our model with AdamW using learning rates $2\mathrm{e}{-5}$ with a linear warmup for the first 6\% of steps. We set the matrix size $N = 42$. The context-based strategy is utilized by default. We tuned the hyperparameters on the development set. We trained on one NVIDIA V100 16GB GPU and evaluated our model with Ign F1, and F1 following \cite{Yao2019DocREDAL}.  

\subsection{Results on the DocRED Dataset}
We compare DocuNet with graph-based models, including GEDA \cite{li2020graph}, LSR~\cite{Nan2020ReasoningWL}, GLRE~\cite{wang2020global} and GAIN~\cite{zeng2020double}, HeterGSAN \cite{xu2020document}; and transformer-based models, including BERT$_{\text{base}}$~\cite{Wang2019FinetuneBF}, BERT-TS$_{\text{base}}$~\cite{Wang2019FinetuneBF}, HIN-BERT$_{\text{base}}$~\cite{tang2020hin}, CorefBERT$_{\text{base}}$~\cite{Ye2020CoreferentialRL}, and ATLOP$_{\text{base}}$ on the DocRED dataset. From the Table \ref{tab::main_results}, we observed that our approach DocuNet-BERT$_{\text{base}}$ obtains better results than ATLOP-BERT$_{\text{base}}$. Moreover, we found that our DocuNet model obtain a new state-of-the-art result  with with RoBERTa-large. As of the IJCAI deadline on 20th of January 2021, \textbf{we held the first position on the CodaLab scoreboard\footnote{\url{https://competitions.codalab.org/competitions/20717\#results}} under the alias \emph{DocuNet}  without external data}\footnote{The SSAN\_ADAPT model leverages pre-training with external distance supervised data.}. 

 \begin{figure*}[h]
  \centering 
  \includegraphics[width=0.8\textwidth]{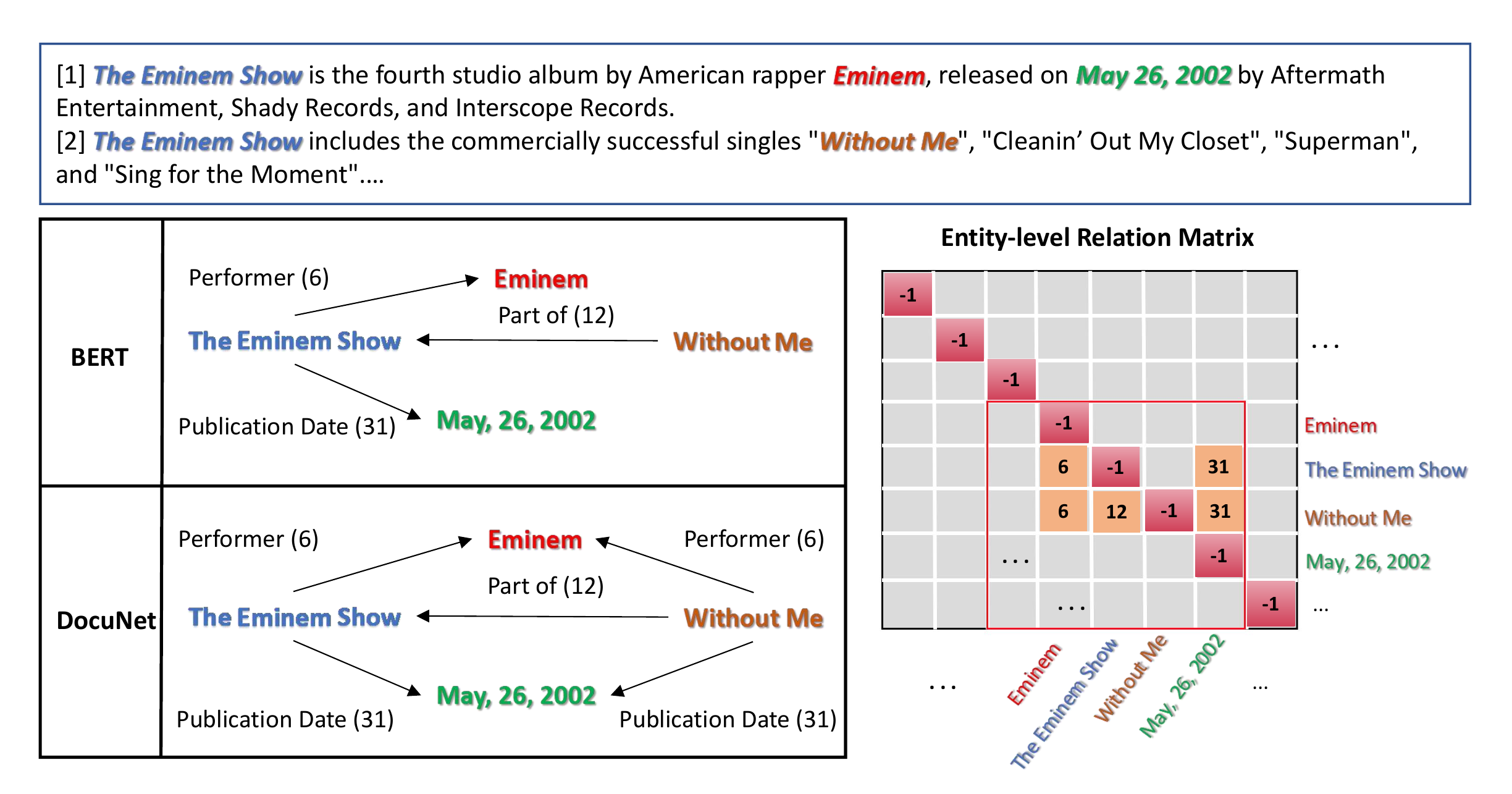} 
  \caption{Case study on our proposed DocuNet and baseline model. The specific number in the figure indicates the corresponding label id.}
  \label{case}
\end{figure*}

\subsection{Results on the Biomedical Datasets}
In the biomedical datasets, we compare DocuNet with lots of baselines including: BRAN \cite{Verga2018SimultaneouslyST}, EoG \cite{Christopoulou2019ConnectingTD}, LSR \cite{Nan2020ReasoningWL}, DHG \cite{zhang2020document}, GLRE \cite{wang2020global} and ATLOP \cite{zhou2020document}. Following ATLOP\cite{zhou2020document}, we utilize the SciBERT \cite{Beltagy2019SciBERTAP} which is pre-trained on the scientific publication corpora. From the Table \ref{tab::bio_result}, we observe model DocuNet-SciBERT$_{\text{base}}$  improved the F1 score by \textbf{6.9\%} and \textbf{1.4\%} on CDR and GDA compared with ATLOP-SciBERT$_{\text{base}}$,

\begin{table}[!t]
\centering
    \begin{tabular}{p{5.2cm}cc}
         \toprule
         \textbf{Model} & Ign $F_1$ & $F_1$ \\
         \midrule
         DocuNet (Context-based) &\textbf{59.86} &\textbf{61.83}  \\
         DocuNet (Similarity-based)&59.04 &60.92 \\
         \midrule
         $w/o$  Balanced Softmax&58.56  &60.51  \\
         $w/o$  U-shaped Segmentation&57.51  &59.65  \\
         \bottomrule
    \end{tabular}
    \caption{Ablation study of DocuNet on DocRED.}
    \label{tab::ablation}
\end{table}

\subsection{Ablation Study}
We conducted an ablation study experiment to validate the effectiveness of different components of our approach. \textbf{DocuNet (Similarity-based)}
 means directly using similarity functions strategy to calculate the correlation between two entities as the input matrix, rather than context-based strategy. \textbf{w/o U-shaped Segmentation} means that our segmentation module is replaced by a feed-forward neural network. \textbf{w/o balanced softmax} refers to the model only with binary cross-entropy loss. From Table \ref{tab::ablation}, we observe that all models have a performance decay without each module, which indicates that both components are beneficial. Besides, we observed that the U-shaped segmentation module and balanced softmax module are most important to model performance and sensitive to $F_1$, leading to a drop of $2.18\%$ and $1.32\%$ in dev $F_1$ score respectively when removed from DocuNet. That reveals that global interdependency among triples captured by our model is effective for document-level RE.
 Moreover, compared with context-based strategy, our approach based on similarity functions strategy drop by 0.84 $F_1$, which illustrates the context-based strategy is advantageous.

\subsection{Case Study}
We follow GAIN~\cite{zeng2020double} to select the same example and  conduct a case study to further illustrate the effectiveness of our model DocuNet compared with the baseline. As shown in Figure \ref{case}, we notice that both BERT$_{base}$ and DocuNet-BERT$_{base}$ can successfully extract the ``part of'' relation between \emph{``Without Me''} and  \emph{``The Eminem Show''}.
However, only our model DocuNet-BERT$_{base}$ is able to deduce that the ``performer'' and ``publication date'' of \emph{``Without Me''} are the same as those of \emph{``The Eminem Show''}, namely, \emph{``Eminem''} and \emph{``May 26, 2002''}, respectively.

Intuitively we can observe that relation extraction mentioned above among those entities requires logical inference across sentences. This interesting observation indicates that our U-shaped segmentation structure over the entity-level relation matrix may implicitly conduct relational reasoning among entities. 
 
\begin{figure}[h]
  \centering 
  \includegraphics[width=0.5\textwidth]{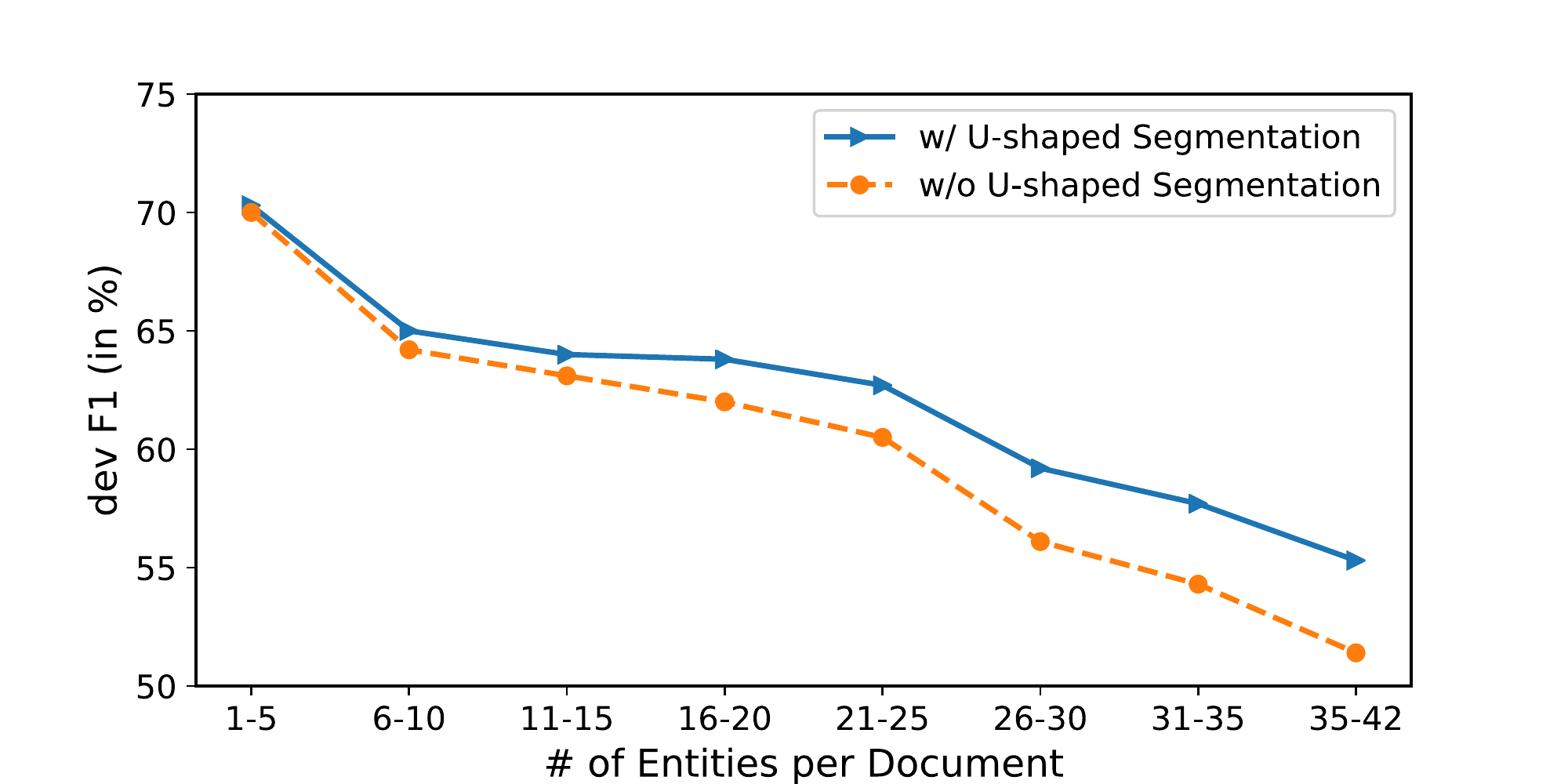} 
  \caption{Dev results in terms of number of entities on  DocRED.}
 \label{plot}	
\end{figure}

\subsection{Analysis}
To assess the effectiveness of DocuNet in modeling global information for multiple entities, we evaluated models respectively trained with or without U-shaped segmentation module on different groups of development set in DocRED, which are divided by the number of entities. From Figure \ref{plot}, we observe that the model w/ U-shaped segmentation module consistently outperforms the model w/o U-shaped segmentation module. We notice that when the number of entities increases, the improvement becomes larger. This indicates that our U-shaped segmentation module can implicitly learn the interdependency among the multiple triples in one context, thus improving the document-level RE performance.

\section{Conclusion and Future Work} 

In this study, we took the first step in formulating document-level RE as a semantic segmentation task and introducing the Document U-shaped Network. Experimental results showed that our model could achieve better performance by capturing local and global information than baselines. We also empirically observe that convolution over entity-entity relation matrix may implicitly conduct relational reasoning among entities. In the future, we plan to apply our approach to other span-level classification tasks, such as aspect-based sentiment analysis and nest named recognition.

\section*{Acknowledgments}
We  want to express gratitude to the anonymous reviewers for their hard work and kind comments. We thank Ning Ding for helpful discussions and feedback on this paper. This work is funded by National Key R\&D Program of China (Funding No. 2018YFB1402800), NSFC91846204.

\bibliographystyle{named}
\bibliography{ijcai21}

\end{document}


\maketitle

\section{Implementation Details}

\subsection{Encoder Module}
Given the document $d=[x_t]_{t=1}^L$, we insert special symbols "$<e>$" and "$</e>$" at the start and end of mentions to mark the entity positions. We leverage the bert-base-cased and roberta-large for the encoder of DocuNet-BERT$_{\text{base}}$ and DocuNet-RoBERTa$_{\text{large}}$ respectively to obtain the embedding of document. Considering some documents are longer than 512, we thus leverage a \emph{dynamic window} to encode whole documents, which average the embeddings of overlapping tokens of different windows with the fixed length of 512 to obtain the final representations.
For each entity  $e_i$ with mentions $\{m_j^i\}_{j=1}^{N_{e_i}}$, we leverage a smooth version of max pooling, namely, logsumexp pooling~\cite{Jia2019DocumentLevelNR}, to obtain the entity embedding. The entity-level relation matrix can be calculated by \emph{similarity-based} method and \emph{context-based} method which are described concretely in our paper. For \emph{similarity-based} method, we directly apply three  kinds of similarity functions to obtain the 3-dims feature vector based on two entities. For \emph{context-based} method, the initial feature vector is 768-dims, which is too large for subsequent modules. Thus, we adopt MLP to compress feature vector information to 3 dimensions. To sum up, we generate the representation of entities with 768-dims and entity-level relation matrix with 3 channels through the Encoder Module.

\subsection{U-shaped Segmentation Module}
This module is formed as a U-shaped structure, which contains two down-sampling blocks and two up-sampling blocks with skip connections. On the one hand, each down-sampling block has two subsequent max pooling and separate convolution modules. Further, the number of channels is doubled in each down-sampling block.
As the module's input is an entity-level relation matrix with 3 channels, the number of channels are respectively [3, 256, 512, 256, 128, 256] in the entire module sequence.

\subsection{Classification Module}
Given the entity pair embedding $\textbf{e}_s$ and $\textbf{e}_o$ with the entity-level relation matrix $Y$, we respectively concatenate the specified feature vector in matrix with the embedding of $\textbf{e}_s$ and $\textbf{e}_o$ , and map them to hidden representations $z$ with a feedforward neural network. Then, we obtain the probability of relation via a bilinear function.

\section{Experiments Details}
\textbf{Our code is available in the supplementary materials for reproducibility}.  We detail the training procedures and hyperparameters for each of the datasets. We utilize Pytorch \cite{paszke2017automatic} to conduct experiments with one NVIDIA V100 16GB GPU.   All optimization was performed with the Adam optimizer with a linear warmup of learning rate over the first 6\% of steps, then linear decay over the remainder of the training. Gradients were clipped if their norm exceeded 1.0, and weight decay on all non-bias parameters was set to 5e-4. The grid search was used for hyperparameter tuning (maximum values bolded below), using five random restarts for each hyperparameter setting for all datasets. Early stopping was performed on the development set. The batch size was 4 in all cases.

\textbf{DocRED.}
The DocRED dataset is available in \url{https://github.com/thunlp/DocRED}. The hyper-parameter search space was:
\begin{itemize}
\item epoch: 30
\item batch size: 5
\item accumulate: 1
\item bert learning rate: [1e-5, 2e-5, \textbf{3e-5}, 4e-5]
\item unet learning rate: [2e-4, 3e-4, \textbf{4e-4}, 5e-4]
\item warmup rate: 0.06
\end{itemize}

\textbf{CDR and GDA.}
 The CDR and GDA datasets can be obtained following the instructions in \url{https://github.com/fenchri/edge-oriented-graph}

\textbf{CDR.}
 The hyper-parameter search space was:
\begin{itemize}
\item epoch: 30
\item batch size: 4
\item accumulate: 1
\item bert learning rate: [1e-5, 2e-5, \textbf{3e-5}, 4e-5]
\item unet learning rate: [2e-4, 3e-4, \textbf{4e-4}, 5e-4]
\item warmup rate: 0.06
\end{itemize}

\textbf{GDA.}
The hyper-parameter search space was:
\begin{itemize}
\item epoch: 10
\item batch size: 4
\item accumulate: 4
\item bert learning rate: [1e-5, 2e-5, \textbf{3e-5}, 4e-5]
\item unet learning rate: [2e-4, \textbf{3e-4}, 4e-4, 5e-4]
\item warmup rate: 0.06
\end{itemize}

\bibliographystyle{named}
\bibliography{ijcai21}